\documentclass[10pt,letterpaper]{article}
\usepackage[top=0.85in,footskip=0.75in]{geometry}
\pdfoutput=1
\usepackage{changepage}
\usepackage[utf8]{inputenc}
\usepackage{textcomp,marvosym}
\usepackage{fixltx2e}
\usepackage{amsmath,amssymb}
\usepackage{nameref,hyperref}
\usepackage[right]{lineno}
\usepackage{microtype}
\DisableLigatures[f]{encoding = *, family = * }
\usepackage{rotating}
\usepackage{todonotes}
\usepackage{nameref}

\DeclareMathOperator*{\argmax}{arg\,max}

\usepackage[aboveskip=1pt,labelfont=bf,labelsep=period,justification=raggedright,singlelinecheck=off]{caption}
\makeatletter
\renewcommand{\@biblabel}[1]{\quad#1.}
\makeatother
\reversemarginpar

\usepackage{lastpage,fancyhdr,graphicx}
\usepackage{epstopdf}
\DeclareFontEncoding{LGR}{}{}
\DeclareTextSymbol{\~}{LGR}{126}
\usepackage{subfig}

\begin{document}
\vspace*{0.35in}

\begin{flushleft}
{\Large
\textbf\newline{Estimating Nonlinear Dynamics with the ConvNet Smoother}
}
\newline
\\
Luca Ambrogioni\textsuperscript{*}, Umut Güçlü, Eric Maris and Marcel A. J. van Gerven
\\
\bigskip
Radboud University, Donders Institute for Brain, Cognition and Behaviour, Nijmegen, The Netherlands 
\\
\bigskip
* l.ambrogioni@donders.ru.nl

\end{flushleft}
\section*{Abstract}
Estimating the state of a dynamical system from a series of noise-corrupted observations is fundamental in many areas of science and engineering. The most well-known method, the Kalman smoother (and the related Kalman filter), relies on assumptions of linearity and Gaussianity that are rarely met in practice. In this paper, we introduced a new dynamical smoothing method that exploits the remarkable capabilities of convolutional neural networks to approximate complex nonlinear functions. The main idea is to generate a training set composed of both latent states and observations from an ensemble of simulators and to train the deep network to recover the former from the latter. Importantly, this method only requires the availability of the simulators and can therefore be applied in situations in which either the latent dynamical model or the observation model cannot be easily expressed in closed form. In our simulation studies, we show that the resulting ConvNet smoother has almost optimal performance in the Gaussian case even when the parameters are unknown. Furthermore, the method can be successfully applied to extremely nonlinear and non-Gaussian systems. Finally, we empirically validate our approach via the analysis of measured brain signals.

\section{Introduction}
Estimating the state of a dynamical system from a finite set of indirect and noisy measurements is a key objective in many fields of science and engineering~\cite{sarkka2013bayesian}. For example, meteorological forecasting requires the estimation of the dynamical state of a series of atmospheric variables from a sparse set of noisy measurements \cite{krishnamurti2016review}. Another example, essential for several modern technologies, is the localization of physical objects from indirect measurements such as radar, sonar or optical recordings \cite{sarkka2013bayesian}. Even the human brain consistently deals with this problem as it has to integrate a heterogeneous series of indirect sensory inputs in its internal representation of the external world \cite{mathys2014uncertainty}. In the rest of the paper we will refer to this class of problems as \emph{dynamical smoothing}, considering the related problem of \emph{dynamical filtering} as a special case (i.e. 0-lag smoothing). 

In this paper, we introduce a new nonlinear smoothing approach that only requires the ability to sample from measurements and dynamical models by leveraging the exceptional flexibility and representational capabilities of deep convolutional neural networks (ConvNets)~\cite{schmidhuber2015deep}. The key idea is to generate synthetic samples of signal and noise from an ensemble of generators in order to train a deep neural network to recover the latent dynamical state from the observations. This allows the use of very realistic models, where the signal and noise structure can be tailored to the specific application without any concern about their analytic tractability. Furthermore, the procedure completely circumvents the problem of model selection and parameter estimation since the training set can be constructed by hierarchically sampling the model and its parameters from an appropriate ensemble. Importantly, since we can generate an arbitrarily large number of training data, we can train arbitrarily complex deep networks without over-fitting. 

\subsection{Related works}
Conventional solutions to the dynamical smoothing problem usually rely on a series of mathematical assumptions on the nature of the signal and the measurement process. For example, when the state dynamic is linear, the measurement model is Gaussian and all the parameters are known, the optimal solution is given by the Kalman smoother (also known as the Rauch–Tung–Striebel smoother)~\cite{briers2010smoothing}. For nonlinear state dynamics and/or a non-Gaussian measurement model, the dynamical smoothing problem can no longer be solved exactly. In these cases a common approximation is the extended Kalman smoother (EKS), which works by linearizing the state dynamics and the measurement model at each time point~\cite{sorenson1985kalman}. A more modern approach is the unscented Kalman smoother (UKS) that approximates the dynamical smoothing distribution by passing a set of selected points (sigma points) through the exact nonlinear functions of the state dynamics and the measurement model~\cite{wan2000unscented}. Unfortunately, these methods may introduce systematic biases in the estimated state and require the availability of both a prior and a likelihood function in closed form. In theory these shortcomings can be overcome by using sampling methods such as particle smoothers~\cite{sarkka2013bayesian}. However, these methods require a large number of samples in order to be reliable and are affected by particle degeneration.

In recent years, several authors pioneered the use of deep neural networks and stochastic optimization on dynamical filtering and smoothing problems. Haarnoja and collaborators introduced the backprop KF, a deterministic filtering method based on a low-dimensional differentiable dynamical model whose input is obtained from high-dimensional measurements through a deep neural network \cite{haarnoja2016backprop}. Importantly, all the parameters of this model can be optimized using stochastic gradient descent. Improvements in stochastic variational inference led to several applications to dynamical filtering and smoothing problems. Using variational methods, Krishnan and collaborators introduced the deep Kalman filters \cite{krishnan2015deep}. In this work, the authors assume that the distribution of the latent states is Gaussian with mean and covariance matrix determined from the previous state through a parameterized deep neural network. The parameters of the network are trained using stochastic backpropagation \cite{rezende2014stochastic}. Similarly, Archer and collaborators introduced the use of stochastic black-box variational inference as a tool for approximating the posterior distribution of nonlinear smoothing problems \cite{archer2015black}. Despite the impressive performance of modern variational inference techniques, these methods share with the more conventional methods such as EKF and UKF the problem of introducing a systematic bias due to the constrains in the model and in the variational distribution. Conversely, the bias of our method can be arbitrarily reduced since neural networks are universal function approximators and we can use a limitless number of training samples.

\section{The model}
A state space model is defined as a dynamic equation that determines how the latent state evolves through time, and a measurement model, that determines the outcome of measurements performed on the latent state. If we assume that the measurement process does not affect the latent state, a general stochastic state model can be expressed as follows
\begin{flalign}
  &x(t) = F_{\theta} \big[x(0),...,x(t-1); \xi_t\big]\\
  &y(t) = G_{\phi} \big[x(0),...,x(t); \zeta_t\big]~,
  \label{eq: state space model}
\end{flalign}
where $x(t)$ is the latent state at time $t$ and $y(t)$ is the measurement at time $t$. The functions $F_{\theta}$ and $G_\phi$ determine the dynamic evolution of the latent state and the measurement process respectively. Since the system is causal and is not disturbed by the measurements, the dynamic function $F_{\theta}$ takes as input the past values of $x$ and a random variable $\xi_t$ that introduces stochasticity in the dynamics. Analogously, the measurement function $G_{\phi}$ takes as inputs the past and present values of $x$ and a random variable $\zeta_t$ that accounts for the randomness in the measurements. 

In a dynamical smoothing problem, we aim to recover the latent state $x$ from the set of measurements $\left\{y(0), ..., y(T)\right\}$, where $T$ is the final time point. Usually, the functional form of $F_{\theta}$ and $G_{\phi}$ is assumed to be known in advance. Nevertheless, these functions typically depend on a set of parameters ($\theta$ and $\phi$ respectively) that need to be inferred from the data.

\subsection{Deterministic smoothing as a regression problem}
We will focus on deterministic smoothing, where the output of the analysis is a point estimate of $x$ instead of a full probability distribution. If we have a training set where both the dynamical state $x$ and the measurements $y$ are observable, then the problem reduces to a simple regression problem where we construct a deterministic mapping $f(y;w) = \hat{x}$ by minimizing a suitable cost function of the form 
\begin{equation}
  C(w) = \sum_j D\big[x(t_j),f\big(y(t_j);w\big)\big]~,
\end{equation}
over the space of some parameters $w$. In this expression, the function $D\big[x,\hat{x}\big]$ is some sensible measure of deviation between the real dynamical state $x$ and our estimate $\hat{x}$.

In most situations the state $x$ is not directly observable. However, in the usual dynamical smoothing setting, we assume to have access to the dynamical model that generated the data and the measurements. In this case, the model can simply be used for generating a synthetic training set. If the dynamical model is not exactly known, we can still construct a training set by sampling from a large parametrized family of plausible models. Clearly, this requires the use of a sophisticated regression model that is able to learn very complicated functional dependencies. To this end, we make use of convolutional neural networks, a regression (and classification) method, which has been shown to achieve state-of-the-art performance in many machine learning tasks~\cite{krizhevsky2012imagenet, he2016deep, oord2016wavenet}.

\subsection{Convolutional neural networks}
In this subsection, we briefly describe the details of the convolutional neural network which was used in our experiments. The network comprised a number of dilated convolution layers~\cite{Yu2015} with 60 one-dimensional kernels of length three and rectified linear units, followed by a fully-connected layer with $n$ one-dimensional kernels of length $n$ where $n$ is the signal length. In most experiments, $n$ was 200.

Dilated convolution layers are similar to regular convolution layers with the exception that successive kernel elements have holes between each other, whose size is determined by a dilation factor. As a result, they ensure that the feature map length remains the same as the receptive field length increases. Note that regular convolution layers can be considered dilated convolution layers with a dilation factor of one.

The dilation factor of the first two layers were one, which doubled after every subsequent layer. The number of convolution layers was chosen to be the largest possible value such that the receptive field length of the last convolution layer was less than the signal length. That is,
\begin{equation}
m = \argmax_x\left(3 + 2\sum_{i = 0}^{x - 2}2^i\right) < n
\end{equation}
where $m$ is the number of convolution layers. In all experiments, $m$ equalled seven.

We initialized the bias terms to zero and the weights to samples drawn from a scaled Gaussian distribution~\cite{He2015}. We used Adam~\cite{Kingma2014} with initial $\alpha$ = 0.001, $\beta_1$ = 0.9, $\beta_2$ = 0.999, $\epsilon = 10^{-8}$ and a mini-batch size of 1500 to train the network for 20,000 epochs by iteratively minimizing a smooth approximation of the Huber loss function~\cite{Huber1964}, called the pseudo-Huber loss function~\cite{Charbonnier1997}:
\begin{equation}
 C(w) = \sum_j \sqrt{1 + \left(x(t_j) - f\big(y(t_j);w\big)\right) ^ 2} - 1 \,.
\end{equation}

\subsection{The ConvNet smoother}
The idea behind the ConvNet smoother is simple. We train a convolution neural network to recover the sequence of simulated dynamical states $\{x(0), ..., x(T)\}$ from the set of simulated measurements $\{y(0), ..., y(T)\}$. The network is trained on simulated data that are sampled using the state evolution function $F_{\theta}$  and the measurement function $G_{\phi}$. If we do not know the parameters $\theta$ and $\phi$ in advance, the training set can still be constructed by randomly sampling $\theta$ and $\phi$ prior to each sample of $\{x(0), ..., x(T)\}$ and $\{y(0), ..., y(T)\}$. In this way, we leave to the network the burden of adapting to the specific parameter setting every time a new series of observations is presented as input. Clearly, if the parameter space is large, we would need a more complex network in order to properly learn the mapping. Fortunately, since we can generate an arbitrarily large number of data points, we can potentially train any complex network without overfitting on the training set.

\section{Experiments}
In the following, we validate the ConvNet smoother in two simulation studies and in an analysis of real brain data acquired using magnetoencephalography (MEG). 

\subsection{Analysis of Gaussian dynamics}
When the latent dynamical state is a Gaussian process (GP) with known covariance function, the optimal smoother (in a least-squares sense) is given by the expected value of a GP regression ~\cite{rasmussen2006gaussian}. Here, we compare the performance of the ConvNet smoother with the expectation of both an exact GP regression (with known covariance and noise parameters) and an optimized GP regression (where the parameters are obtained by maximizing the marginal likelihood of the model given the data). The ConvNet was trained with samples drawn from a GP equipped with a squared exponential covariance function~\cite{rasmussen2006gaussian}. For each sample, the length scale and the amplitude parameters were sampled from a log-normal distribution (length scale: $\mu = -1.9$, $\sigma = 0.8$; amplitude: $\mu = 0$, $\sigma = 0.5$). Hence, the ConvNet smoother is effectively trained on a large family of GPs. These samples where then corrupted with Gaussian white noise whose standard deviation was itself sampled from a log-normal distribution ($\mu = -0.9$, $\sigma = 0.5$). The network was trained to recover the ground truth function from the noisy data. 

\begin{figure}[!t]
	\centering
    	\includegraphics[width=1\textwidth] {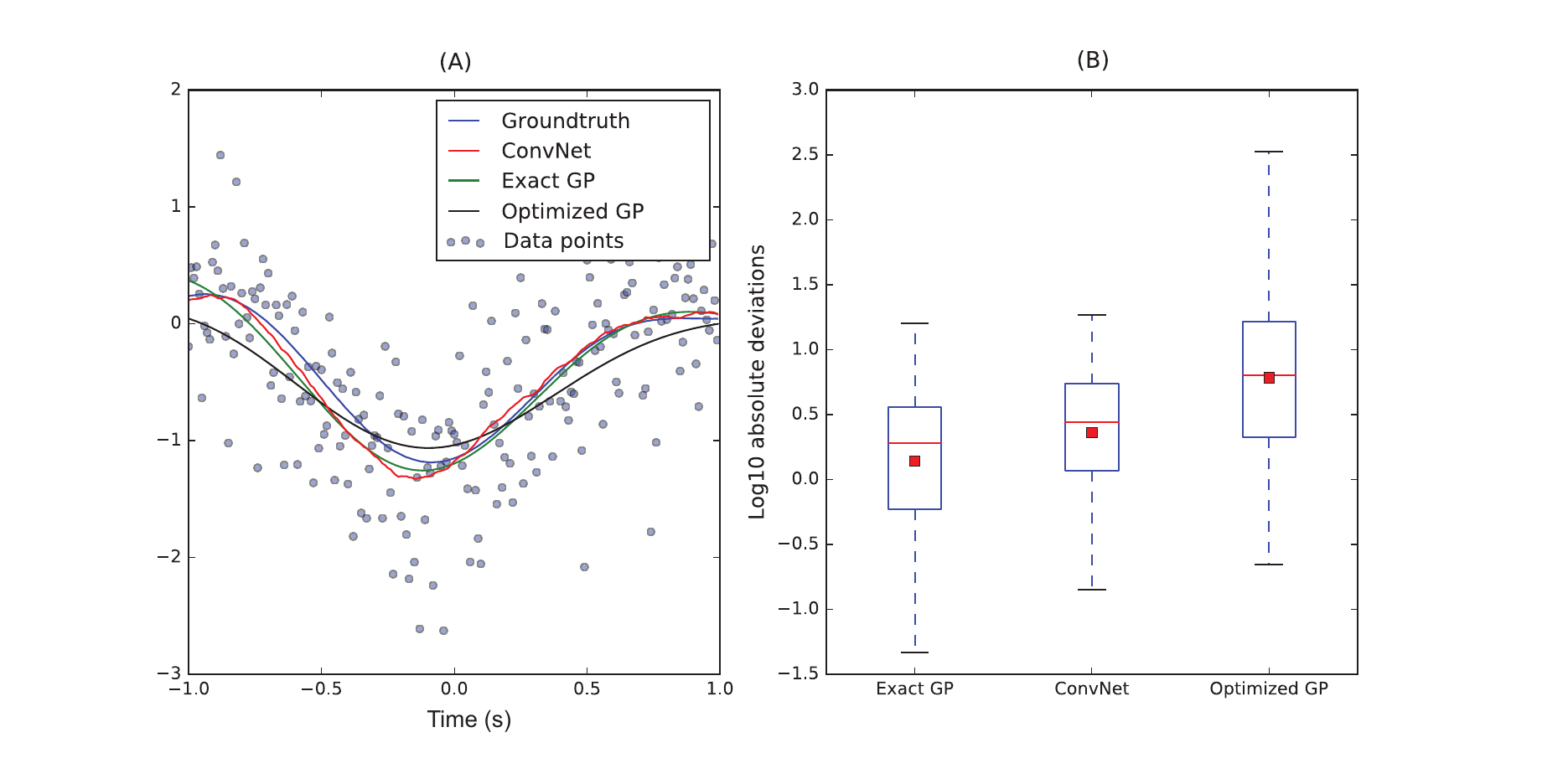}
	\caption{Analysis of Gaussian signals. A) Analysis of an example signal. B) Performance of ConvNet, exact GP and optimized GP. The boxplot shows the absolute deviation between reconstructed signal. The red lines show the median while the red squares show the mean.}
	\label{figure 1}
\end{figure}

Figure~\ref{figure 1} shows the results on a test set comprised of $200$ trials. Panel A shows an example trial. The estimate obtained using the ConvNet smoother is less smooth than the GP estimates but it does a good job at tracking the ground truth signal. Panel B shows the absolute deviations of the ConvNet, exact GP and optimized GP models. The performance of the ConvNet is only slighly worse than the (optimal) exact GP. Furthermore, the ConvNet significantly outperforms the GP optimized by maximal likelihood.

\subsection{Analysis of nonlinear dynamics}
In this section, we show that the ConvNet smoother can be used in complex situations where neither the exact likelihood or the exact prior can be easily expressed in closed form. This freedom allows to train the network on a very general noise model that is likely to approximate the real noise structure of the measured data as a special case.

As dynamical model, we used the following stochastic anharmonic oscillator equation:
\begin{equation}
\frac{d^2x}{dt^2} = -\omega_0^2 x - \beta \frac{dx}{dt} + k_2 x^2 + k_3 x^3 ~ + \xi(t)\,,
\label{eq: nonlinear SDE}
\end{equation}
where $\omega_0$ is the asymptotic undampened angular frequency for small oscillations, $\beta$ is the damping coefficient and both $k_2$ and $k_3$ regulate the deviation from harmonicity. The term $\xi(t)$ is additive Gaussian white noise and introduces stochasticity to the dynamics. We discretized the stochastic dynamical process using the Euler-Maruyama scheme with integration step equal to $0.01$ seconds. The parameters of the dynamical model were kept fixed ($\omega_0 = 5$, $\beta = 0.2$, $k_2 = 15$, $k_3 = -0.5$). This procedure gave a total of $N = 200$ time points for each simulated trial. The experiment is divided in two parts. In the first, we used a Gaussian observation model and we compared our method with existing dynamical smoothing techniques. In the second, we used a more complex parameterized observation model in order to demonstrate the flexibility of the ConvNet method.

\subsubsection{Gaussian observation model}
In this first part of the experiment, we use an observation model where the measurements are corrupted by Gaussian white noise. The standard deviation  of the measurement noise was equal to $20$. We generated a total of 49900 training pairs $\left(\{x(t_1),...,x(t_N)\},\{y(t_1),...,y(t_N)\}\right)$ and 1000 test pairs. We compared the ConvNet smoother with EKS and UKS. 

\begin{figure}[!t]
	\centering
    	\includegraphics[width=1\textwidth] {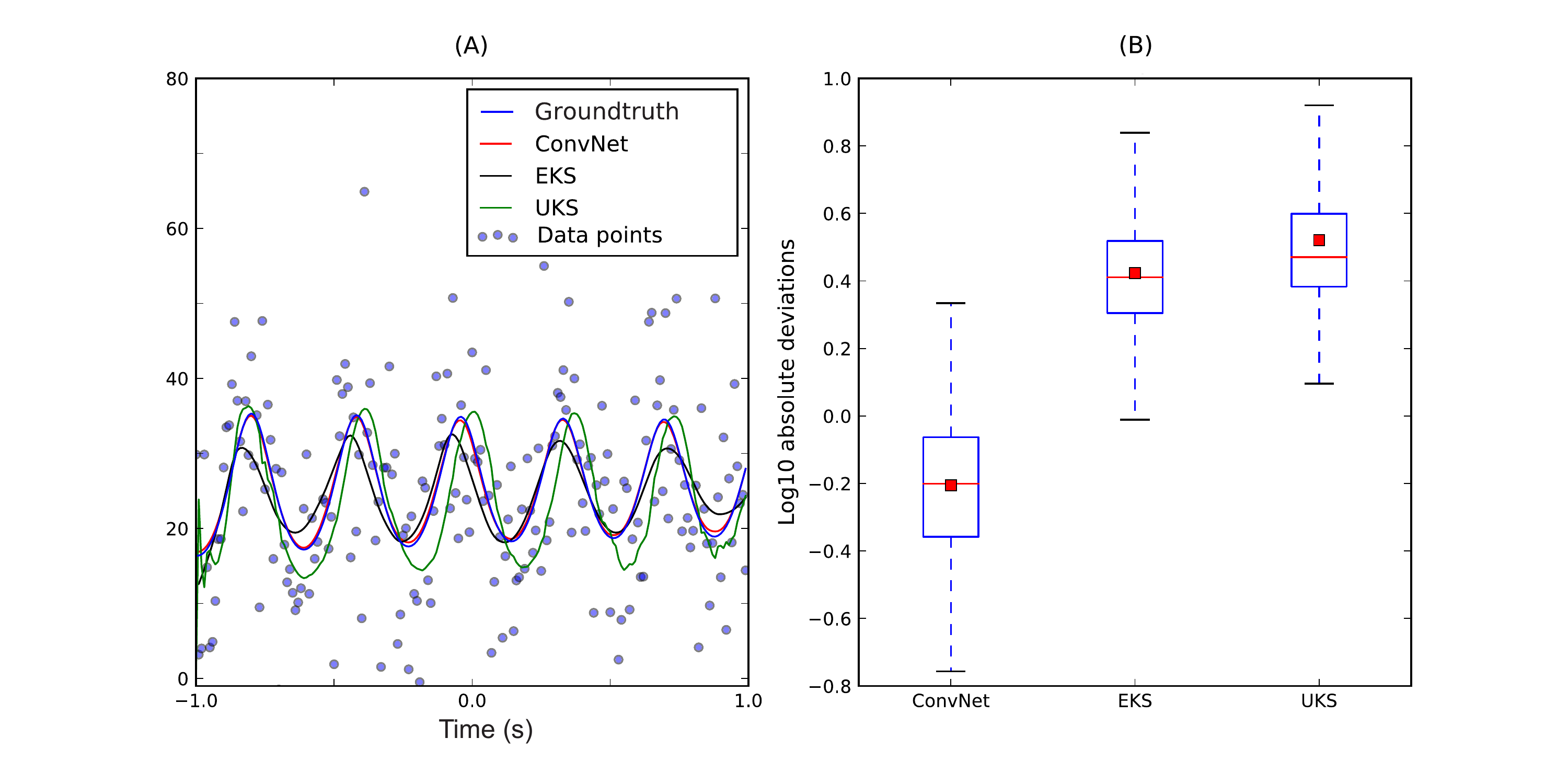}
	\caption{Analysis of nonlinear signals corrupted by conditionally independent noise. A) Analysis of an example signal. B) Performance of ConvNet, EKF and UKF. The boxplot shows the log10 of the absolute deviation between reconstructed signal. The red lines show the median while the red squares show the mean.}
	\label{figure 2}
\end{figure}

Figure~\ref{figure 2}, Panel A shows the latent state of an example trial recovered using the ConvNet smoother, EKS and UKS. The absolute deviations between the recovered latent state and the ground truth signal are shown in Panel B. From the boxplots, it is clear that in this experiment the ConvNet greatly outperforms the other methods. 

\subsubsection{Conditionally dependent observation model}
In this experiment, we use a complex observation model where the measurements are not statistically independent given the latent state. In these situations, EKS and UKS cannot be directly applied. The observations $y(t)$ were obtained as follows:
\begin{equation}
y(t) = x(t) + \eta t + \gamma(t) + \zeta(t)\,,~
\label{eq: observation model}
\end{equation}
where $\eta t$ is a linear trend with slope $\eta$ sampled at random from a normal distribution with zero mean and standard deviation equal to $10$; $\gamma(t)$ is a pure jump process with exponential inter-jump interval (mean equal to $0.5$ s) and scaled Cauchy jump size (scale equal to 1.5); $\zeta(t)$ is Student-t white noise with scale sampled from a gamma distribution (with scale equal to $0.3$, shape equal to $2$) and degrees of freedom sampled from a uniform distribution over the integers from $2$ to $21$. All the parameters of the noise model were sampled every time a new trial was generated. We generated a total of 99900 training pairs $\left(\{x(t_1),...,x(t_N)\},\{y(t_1),...,y(t_N)\}\right)$ and 1000 test pairs.

\begin{figure}[!t]
	\centering
    	\includegraphics[width=1\textwidth] {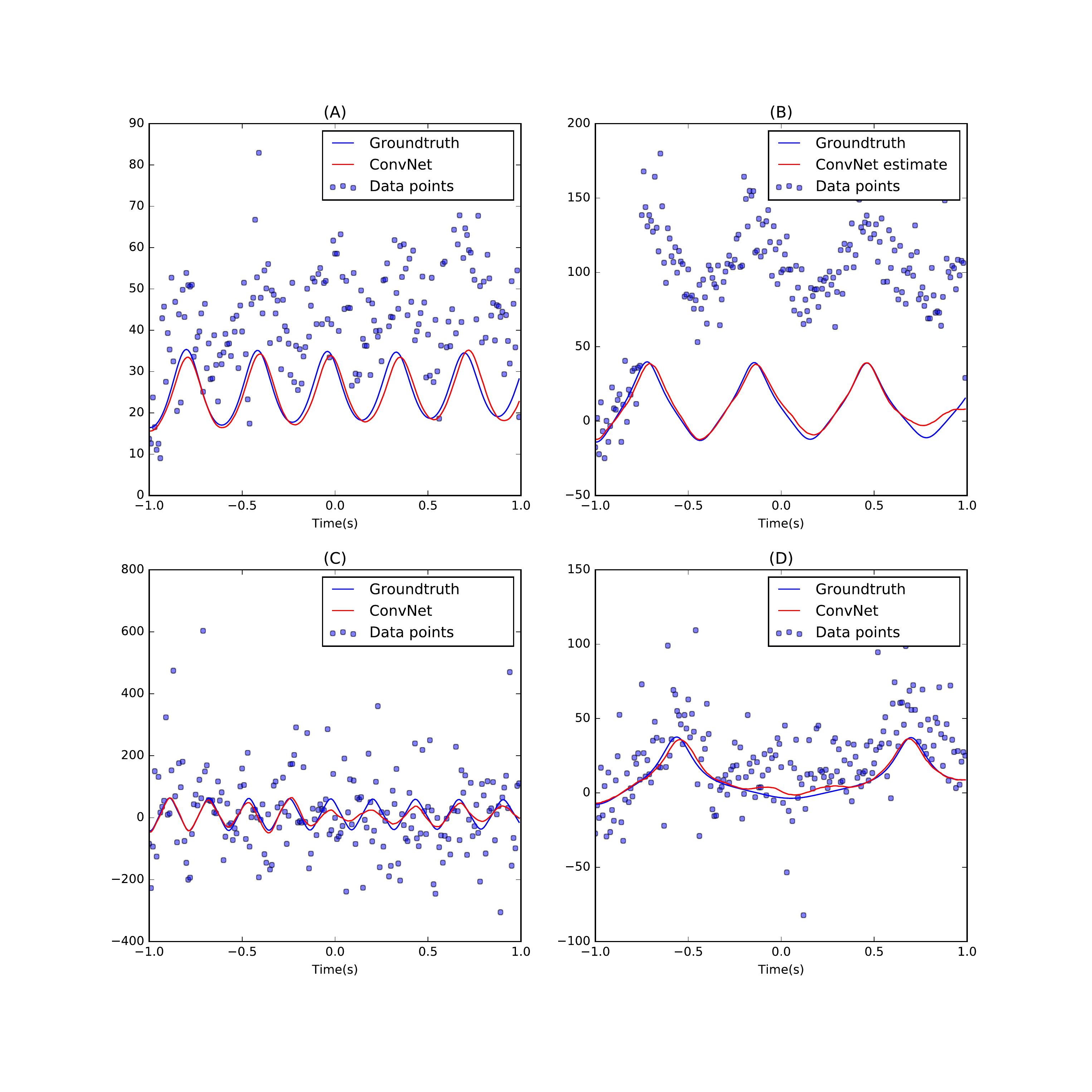}
	\caption{Analysis of nonlinear signals corrupted by conditionally dependent noise. Panels A to D show the input data points (blue dots), ground truth signal (blue line) and ConvNet estimates (red line) for four example trials.}
	\label{figure 3}
\end{figure}

Figure~\ref{figure 3}, Panels A--D shows the results in the test set for four example trials. We can see that these trials are characterized by highly heterogeneous waveforms of the latent dynamical process and very variable noise structure. Visually, the method seems to maintain high performance for a wide range of signal and noise characteristics. For example, Panel~B shows a trial with a large discontinuous jump while Panel~C shows a trial with very pronounced outliers. The median log10 absolute deviation of the model output from the ground truth dynamical signal was $0.32$ while the upper and lower quantiles were $-0.51$ and $1.38$ respectively.

\subsection{Analysis of brain oscillations}
As a final example, we used the ConvNet smoother for recovering 10 Hz brain oscillations (the so-called alpha rhythm; see \cite{cantero2002human}) from MEG measurements. While several neural field equations have been proposed \cite{deco2008dynamic}, so far there is not a universally accepted dynamical model of cortical electromagnetic activity. Therefore, for this application, we generated the dynamical latent state using an empirical procedure that is meant to capture the qualitative features of alpha oscillations without resorting to an explicit equation of motion. The idea is to sample from a sufficiently large family of signal and noise models, in order to capture the observed data as special case. After training, the ConvNet should be able to recognize the appropriate signal and noise characteristics directly from the input time series.

In our example, the oscillatory waveform was generated as follows:
\begin{equation}
x(t) = \mathcal{A}(t) f\bigg(\cos\big(\omega(t) t + \phi_0\big)\bigg)~\,,
\label{eq: oscillations model}
\end{equation}
where the envelope $\mathcal{A}(t)$ and the angular frequency parameter $\omega(t)$ are sampled from a GP and the initial phase $\phi_0$ is sampled from a uniform distribution.
Furthermore, the nonlinear function $f(a)$ has the following form:
\begin{equation}
f(a) = w_1 a + w_2 a^2 + w_3 a^3 + w_4 a^4 + w_5 a^5~,
\label{eq: nonlinear functions}
\end{equation}
where the Taylor coefficients $w_1,w_3$ and $w_5$ were sampled from truncated t distributions (df = 3, from $0$ to $\infty$) and the coefficients $w_2$ and $w_4$ were sampled from t distributions (df = 3). This allows to generate synthetic alpha oscillations with variable waveform, amplitude and peak frequency. 

\begin{figure}[!t]
	\centering
    	\includegraphics[width=1\textwidth] {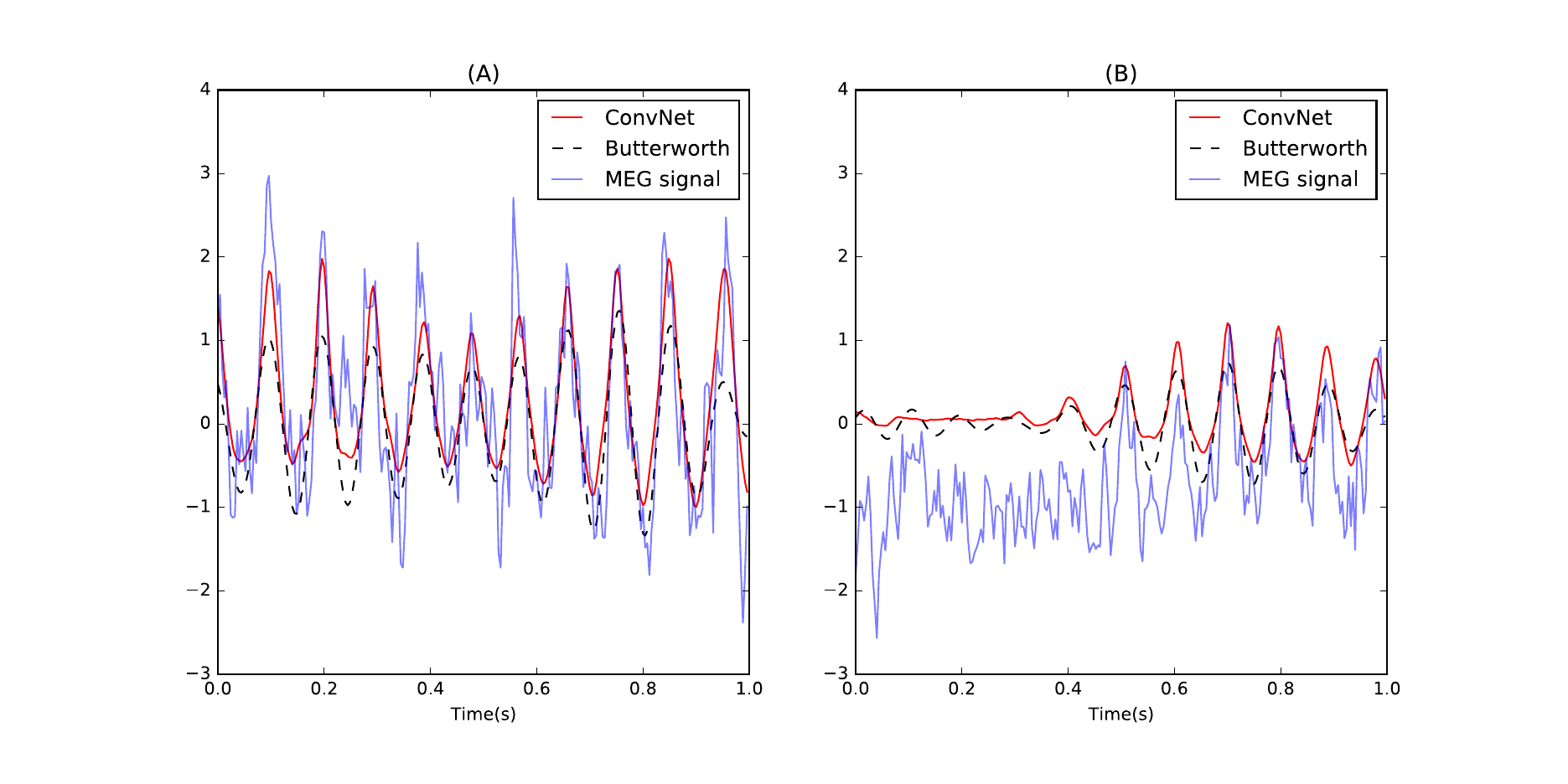}
	\caption{Analysis of brain oscillations. Panels A-B show the raw MEG signal (blue line) and the estimate of the alpha oscillations time series obtained using Butterworth filter (dashed line) and ConvNet smoother (red line) for two example trials.}
	\label{figure 4}
\end{figure}

The network was trained on the synthetic data set and then applied to a resting-state MEG data set. The experimental procedure is described in~\cite{ambrogioni2016complex}. We compared the resulting estimate with the estimate obtained by applying a band-pass Butterworth (two-pass, 4th order, from 8 Hz to 12 Hz) filter to the MEG data. Figure~\ref{figure 4} shows the result in two example trials. In Panel A, the oscillatory alpha activity is very prominent across the whole trial. Note that the ConvNet smoother is able to recover the highly anharmonic waveform without introducing a substantial amount of noise. In Panel B, we can see that the oscillatory activity is absent in the first half of the trial and becomes prominent in the second half. Importantly, the ConvNet smoother almost completely suppresses the oscillatory response in the first part while the linear filter exhibits a low amplitude oscillation.  

\section{Conclusions}
In this paper, we introduced the use of deep convolution neural networks trained on synthetic data for nonlinear smoothing. The ConvNet smoother requires no analytic work by the practitioner besides the design of an appropriate ensemble of signal and noise simulators. Importantly, imperfect prior knowledge about the signal and the noise model is compensated by the remarkable capacity of deep convolutional neural networks to recognize patterns in the data. 

Several improvements are possible. First, the model can easily be used for forecasting by training the network with an initial segment of noisy time series as input and the full noise-free time series as output. This kind of application could have a major impact given the importance of ensemble forecasting methods in fields such as meteorology~\cite{krishnamurti2016review}. Second, the method can be adapted to online filtering by using an autoregressive convolutional architecture where the filter kernels only have access to previous time points~\cite{uria2016neural}. Third, the uncertainty over the latent dynamical signal can be estimated either by drop-out~\cite{gal2015dropout} or by using a conditional density estimation neural network for estimating the full conditional distribution of the dynamical state given the data~\cite{williams1996using}. This latter approach may be considered as an application of the recently introduced framework for Bayesian conditional density estimation~\cite{papamakarios2016fast}.


\bibliography{ConvNet_smoother.bib}

\end{document}